\documentclass[sigconf]{acmart}
\AtBeginDocument{%
  \providecommand\BibTeX{{%
    \normalfont B\kern-0.5em{\scshape i\kern-0.25em b}\kern-0.8em\TeX}}}

\copyrightyear{2024}
\acmYear{2024}
\setcopyright{acmlicensed}
\acmConference[KDD '24] {Proceedings of the 30th ACM SIGKDD Conference on Knowledge Discovery and Data Mining }{August 25--29, 2024}{Barcelona, Spain.}
\acmBooktitle{Proceedings of the 30th ACM SIGKDD Conference on Knowledge Discovery and Data Mining (KDD '24), August 25--29, 2024, Barcelona, Spain}
\acmISBN{979-8-4007-0490-1/24/08}
\acmDOI{10.1145/3637528.3671660}

\usepackage[ruled]{algorithm2e}
\usepackage{subfigure}
\usepackage{makecell}
\usepackage{enumerate}

\newtheorem{problem}{Problem}
\newtheorem{definition}{Definition}
\newtheorem{example}{Example}

\newcommand{\eg}{\emph{e.g.},\xspace}
\newcommand{\ie}{\emph{i.e.},\xspace}

\newcommand{\eat}[1]{}

\begin{document}

\title{HiFGL: A Hierarchical Framework for Cross-silo Cross-device Federated Graph Learning}

\author{Zhuoning Guo}
\email{zhuoning.guo@connect.ust.hk}
\affiliation{%
  \institution{The Hong Kong University of Science and Technology (Guangzhou), The Hong Kong University of Science and Technology}
  \country{}
}

\author{Duanyi Yao}
\email{dyao@connect.ust.hk}
\affiliation{%
  \institution{The Hong Kong University of Science and Technology}
  \country{}
}

\author{Qiang Yang}
\email{qyang@cse.ust.hk}
\affiliation{%
  \institution{The Hong Kong University of Science and Technology}
  \country{}
}

\author{Hao Liu$^{*}$}
\email{liuh@ust.hk}
\affiliation{%
  \institution{The Hong Kong University of Science and Technology (Guangzhou), The Hong Kong University of Science and Technology}
  \country{}
}

\thanks{$^{*}$~Corresponding author.}

\renewcommand{\shortauthors}{Guo, et al.}

\begin{abstract}
Federated Graph Learning (FGL) has emerged as a promising way to learn high-quality representations from distributed graph data with privacy preservation.
Despite considerable efforts have been made for FGL under either cross-device or cross-silo paradigm, how to effectively capture graph knowledge in a more complicated cross-silo cross-device environment remains an under-explored problem.
However, this task is challenging because of the inherent hierarchy and heterogeneity of decentralized clients, diversified privacy constraints in different clients, and the cross-client graph integrity requirement.
To this end, in this paper, we propose a Hierarchical Federated Graph Learning (HiFGL) framework for cross-silo cross-device FGL.
Specifically, we devise a unified hierarchical architecture to safeguard federated GNN training on heterogeneous clients while ensuring graph integrity.
Moreover, we propose a Secret Message Passing (SecMP) scheme to shield unauthorized access to subgraph-level and node-level sensitive information simultaneously.
Theoretical analysis proves that HiFGL achieves multi-level privacy preservation with complexity guarantees.
Extensive experiments on real-world datasets validate the superiority of the proposed framework against several baselines.
Furthermore, HiFGL's versatile nature allows for its application in either solely cross-silo or cross-device settings, further broadening its utility in real-world FGL applications.
\end{abstract}





\maketitle

\section{Introduction}
Federated Learning (FL) has emerged as a transformative approach by enabling multiple parties to contribute to a shared machine learning model without the need for direct data exchange~\cite{QiangYang2019FederatedML,xie2023federatedscope}.
Along this line, Federated Graph Learning~(FGL) has been proposed to collaboratively train the Graph Neural Network~(GNN) to extract distributed knowledge from interconnected subgraphs held privately by each party.
Recently, FGL has been adopted to a wide range of application domains, such as finance~\cite{ToyotaroSuzumura2019TowardsFG}, recommender system~\cite{ZhiweiLiu2022FederatedSR}, and transportation~\cite{XiaomingYuan2022FedSTNGR}.

\begin{figure*}
\setlength{\belowcaptionskip}{-0.3cm}
    \centering
    \includegraphics[width=\linewidth]{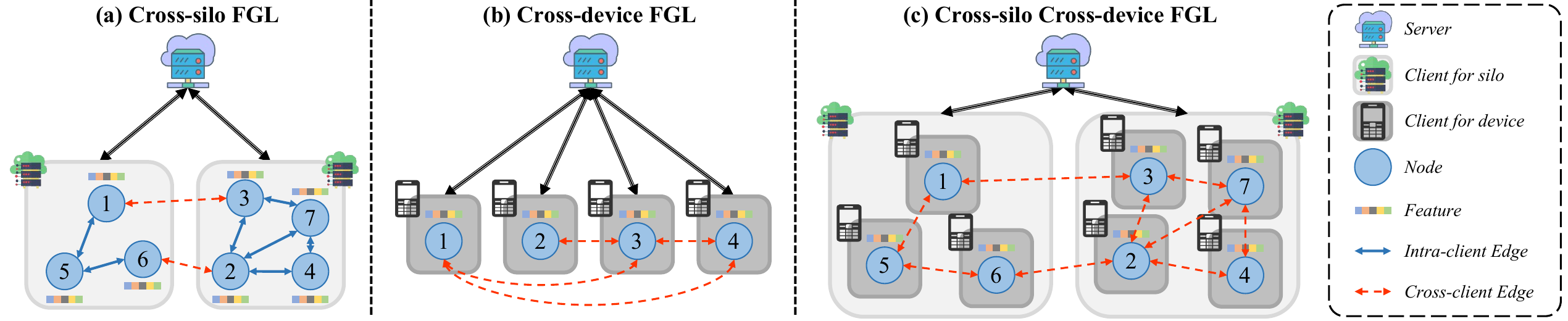}
    \caption{Illustration of three FGL paradigms: cross-silo, cross-device, and cross-silo cross-device FGL.}
    \label{fig:fgl_paradigms}
    
\end{figure*}

Existing FGL approaches predominantly revolve around two paradigms~\cite{ChaoHuang2022CrossSiloFL}.
\textit{1)~Cross-silo FGL}~\cite{zhang2021subgraph,baek2023personalized,wang2022graphfl}, as illustrated in Figure~\ref{fig:fgl_paradigms}~(a), formulates data silos as clients, each of which possessing a subgraph consists of nodes and connected edges. This paradigm is particularly relevant for institutions wishing to maintain private subgraphs while contributing to a global graph structure, such as cross-platform recommendation for E-commerce~\cite{lin2019cross}.
\textit{2)~Cross-device FGL}~\cite{ChuizhengMeng2021CrossNodeFG,wu2022federated,QiyingPan12022FedWalkCE}, as shown in Figure~\ref{fig:fgl_paradigms}~(b), regards each device as a client, which holding a node and its associated edges. It is more suitable for scenarios where numerous devices maintain private connections within a global graph, e.g., user-centric social networks~\cite{qu2023semi}.
Table~\ref{tab:related_works} summarizes the privacy, utility, and efficiency tradeoff of existing FGL approaches.

Despite the success of the above two paradigms in unleashing the power of isolated graph data, none of them can be directly adopted to fully address the complexities of a mixed cross-silo cross-device environment, as demonstrated in Figure~\ref{fig:fgl_paradigms}~(c).
In real-world scenarios, institutions and users may have common privacy concerns but with varying privacy and utility requirements~\cite{XiaojinZhang2022TradingOP,XiaojinZhang2023NoFL}.
We illustrate the use case of FGL under the cross-silo cross-device setting via the following real-world application.

\begin{example}\label{exa:fin}
\textbf{Anomaly detection in financial transactions.}. 
In this scenario, banks oversee customer transactions, forming a federated graph where each bank controls a subgraph of accounts (i.e., nodes) and transactions (i.e., edges).
The collaborative analysis of the cross-bank federated graph can enhance the anomaly detection task. 
However, the transaction records are confidential, and regulations forbid the banks to disclose the subgraph structure.
Meanwhile, customers are also unwilling to expose their sensitive information to institutions to avoid personal privacy leakage.
\end{example}

To bridge the gap, we investigate the cross-silo cross-device federated graph learning problem, where institutions and customer devices collaboratively optimize a GNN model under their diverged privacy constraints.
However, three major technical challenges arise.
\textbf{1)~Inherent hierarchy and heterogeneity of decentralized clients.}
The participants in cross-silo cross-device FGL naturally form a hierarchy, where an institution client may have a more comprehensive local structure of devices, and devices may preserve their local sensitive features.
Besides, the cross-silo cross-device FGL involves clients with varying computational capabilities. Such hierarchy and heterogeneity pose significant challenges in designing a unified FGL framework that can effectively operate across a varied landscape.
\textbf{2)~Diversified privacy constraints in different clients.}
As depicted in Table~\ref{tab:related_works}, privacy concerns in FGL vary across different types of clients.
In a word, cross-silo FGL focuses more on subgraph-level privacy~(i.e., the privacy of structures), while cross-device FGL weighs more on node-level privacy~(i.e., the privacy of features).
The varied privacy requirements necessitate a flexible approach that can adapt to the specific needs of different types of clients while ensuring the overall utility of the federated graph learning process.
\textbf{3)~Cross-client graph integrity.}
In FGL, each client contributes to a portion of the overall graph. Maintaining the integrity of graph data across multiple clients is critical to the utility of the joint model.
However, it is a non-trivial task to protect the cross-client graph information without sacrificing the model performance.
For example, FedSage+~\cite{zhang2021subgraph} protects cross-silo structures by generatively approximating edges across local subgraphs, while Glint~\cite{TaoLiu2021GlintDF} chooses to expose node embeddings to guarantee model utility. 
It is challenging to preserve graph privacy without sacrificing graph integrity. 

\begin{table*}[t]
\setlength{\abovecaptionskip}{0.1cm}
  \centering
  \caption{Comparison of typical FGL works and HiFGL in the dimension of \textit{Multi-level Privacy Preservation} (including \textit{Subgraph-level} and \textit{Node-level}), \textit{Information Integrity}, and \textit{Efficiency}. We evaluate properties by \textit{Low}, \textit{Medium}, or \textit{High} by considering how much a framework satisfies requirements relatively among baselines.
  % Full table is presented in Appendix~\ref{sec:appendix_related_works}.
  }
  \begin{tabular}{c|ccccc}
      \toprule
      Model & FGL Paradigm & \multicolumn{2}{c}{Multi-level Privacy Preservation} & Information Integrity & Efficiency \\
      & & \textit{Subgraph-level} & \textit{Node-level} & & \\
      \midrule
      GCN~\cite{kipf2016semi}+FedAvg~\cite{mcmahan2017communication} & Cross-silo & Medium & Low & Medium & High\\
      FedSage+~\cite{zhang2021subgraph} & Cross-silo & Medium & Low & Medium & Low \\
      FedPUB~\cite{baek2023personalized} & Cross-silo & Medium & Low & Medium & Medium \\
      GraphFL~\cite{wang2022graphfl} & Cross-silo & Medium & Low & Medium & Medium \\
      FedGraph~\cite{FahaoChen2022FedGraphFG} & Cross-silo & High & Low & High & Low \\
      Glint~\cite{TaoLiu2021GlintDF} & Cross-silo & Low & High & High & Medium \\
      PPSGCN~\cite{BinchiZhang2022PPSGCNAP} & Cross-silo & Medium & Low & High & Low \\
      FedCog~\cite{lei2023federated} & Cross-silo & Medium & Low & High & Low \\
      CNFGNN~\cite{ChuizhengMeng2021CrossNodeFG} & Cross-device & - & Medium & High & Medium \\
      FedPerGNN~\cite{wu2022federated} & Cross-device & - & High & High & Low \\
      FedWalk~\cite{QiyingPan12022FedWalkCE} & Cross-device & - & High & High & Medium \\
      Lumos~\cite{pan2023lumos} & Cross-device & - & High & High & Low \\
      SemiDFEGL~\cite{qu2023semi} & Cross-device & - & High & High & Medium \\
      \midrule
      \textbf{HiFGL~(Ours)} & Cross-silo Cross-device (Both) & High & High & High & Medium\\
      \bottomrule
  \end{tabular}
  \label{tab:related_works}
\end{table*}

In this paper, we propose a unified cross-silo cross-device framework, \textit{\textbf{Hi}erarchical \textbf{F}ederated \textbf{G}raph \textbf{L}earning}~\textbf{(HiFGL)}, for multi-level privacy preservation (i.e., both subgraph-level and node-level) without sacrificing graph information integrity. 
Specifically, we first construct a hierarchical architecture comprising three key components: device-client, silo-client, and server. The hierarchical architecture enables federated graph learning with the flexibility of applying diversified privacy preservation strategies on different types of clients.
Moreover, we propose a  \textit{Secret Message Passing} (\textit{SecMP}) scheme for multi-level privacy protection. In particular, a Neighbor-Agnostic Aggregation protocol and a Hierarchical Lagrangian Embedding protocol are proposed to reduce subgraph-level and node-level privacy leakage, respectively.
Furthermore, a tailored resource-efficient optimization algorithm is introduced for the unified framework.
Notably, HiFGL can also be applied in solely cross-silo or cross-device scenarios, and is compatible with diverse FL algorithms~(\eg FedAvg~\cite{mcmahan2017communication}, FedProx~\cite{li2020federated}) and GNN variants~(\eg GCN~\cite{kipf2016semi}, GraphSage~\cite{hamilton2017inductive}), further broaden its utility in real-world FGL applications.

The main contributions of our work are listed as follows:
1)~To our knowledge, HiFGL is the first framework tailored for the cross-silo cross-device federated graph learning problem, which has rarely been studied before.
2)~We construct a hierarchical architecture to enable flexible privacy-preservation of heterogeneous clients without sacrificing graph integrity.
3)~We propose a secret message passing scheme to simultaneously safeguard subgraph-level and node-level privacy against semi-honest adversaries during collaborative GNN training.
4)~We theoretically analyze the privacy and complexity of HiFGL. The results show that our methods not only preserve subgraph-level and node-level privacy but also achieve information integrity with guaranteed complexity.
5)~We conduct extensive experiments on real-world graph datasets, and the results demonstrate that HiFGL outperforms state-of-the-art baselines in learning more effective GNN models.

\section{Related Work}\label{sec:related_work}
In this section, we review state-of-the-art federated graph learning approaches, including two paradigms: cross-silo and cross-device.

\subsection{Cross-silo Federated Graph Learning}
A common scenario of cross-silo FGL is that institutions collaboratively learn models while keeping local data private. Besides various GNN modules for improving effectiveness, existing cross-silo FGL works have different strategies for processing graph structure data.

The first strategy is to drop the cross-client edges to prevent data leakage across clients.
An intuitive way is that each client holds a GNN~(\eg GCN~\cite{kipf2016semi}) and a subgraph without cross-client edges and trains a global model through FedAvg~\cite{mcmahan2017communication}, a custom FL framework.
In addition, FedSage+~\cite{zhang2021subgraph} generates local nodes' neighbor features to offset the ignorance of cross-client edges due to subgraph-level privacy preservation, which improves predicting ability with significant extra training costs.
FedPUB~\cite{baek2023personalized} focuses on the prediction improvement via personalized masked graph convolutional network, where their pairwise similarity is measured by subgraph representation.
GraphFL~\cite{wang2022graphfl} solves the semi-supervised graph learning problem on federated unconnected subgraphs, where node label domains vary and are not identically distributed.
Another strategy is to maintain cross-client edges either stored by the server or clients. In this way, we have to face either less practicability or more privacy leakage.
For example, FedGraph~\cite{FahaoChen2022FedGraphFG} uses a central server to keep cross-client edges and federally trains GNNs with graph sampling through huge communication for expanding neighbors between clients and the server. Unfortunately, the framework is not practical in the real world~\cite{RuiLiu2022FederatedGN}.
Glint~\cite{TaoLiu2021GlintDF} decentralized trains graph convolutional networks, where nodes are fully aware of cross-client neighbors.
PPSGCN~\cite{BinchiZhang2022PPSGCNAP} leverages the graph sampling method to enhance efficiency and scalability, which hides node information but exposes nodes across clients.
FedCog~\cite{lei2023federated} decouples subgraphs according to intra- or cross-client edges to construct a border graph for each client, which is a bipartite graph between internal nodes and external nodes, with two separated graph learning operations.

We propose quantifying subgraph-level privacy leakage in Appendix~\ref{sec:appendix_subgraph_privacy}. The results demonstrate the unsolved issue of balancing subgraph-level privacy protection and cross-client graph integrity. Our approach aims to achieve two objectives simultaneously.

\subsection{Cross-device Federated Graph Learning}
Cross-device FGL assumes users hold private data and learn models without accessing private data.
To name a few, CNFGNN~\cite{ChuizhengMeng2021CrossNodeFG} combines the spatiotemporal GNN model with FL, where the data storage scheme hides the original information of nodes but exposes hidden states and neighboring nodes.
FedPerGNN~\cite{wu2022federated} applies GNNs for recommender systems where a user keeps a local user-item subgraph. A privacy-preserving graph expansion method anonymously acquires neighbor information with high communication costs.
FedWalk~\cite{QiyingPan12022FedWalkCE} adopts the random walk algorithm for federated graph node embedding learning with node-level visibility for covering raw graph information.
Lumos~\cite{pan2023lumos} utilizes local differential privacy and zero-knowledge protocol to protect node features' and degrees' privacy among decentralized devices.
SemiDFEGL~\cite{qu2023semi} collaborates ego graph-corresponded devices via a peer-to-peer manner for scalability improvement and communication reduction on recommendation tasks.
In summary, existing works either utilize inefficient privacy-preserving schemes to align pairwise relationships or rely on a server with extreme representation exposure to achieve cross-client edge integrity.
Our framework constructs a three-level architecture with multiple privacy preservation protocols to achieve effective and secure FGL.

\section{Preliminaries}

\subsection{Federated Graph Definition}
In FGL, graphs are required to be distributively stored with privacy preservation. We focus on the scenario where a complete graph consists of subgraphs without overlapped nodes and nodes have their features and edges~\cite{zhang2021subgraph}.
Let $v$ denote a node associated with a multi-dimensional feature vector $h_v$, an edge $e = (v_i, v_j)$ is defined as a directed linkage from $v_i$ to $v_j$. A subgraph is defined as $\mathbb{G}_s=(\mathbb{V}_s,\mathbb{E}_s)$, where $\mathbb{V}_s$ is the node set and $\mathbb{E}_s$ is the edge set. Note $\exists v_j \notin \mathbb{V}_s, e = (v_i, v_j) \in \mathbb{E}_s$, \ie a node in a subgraph can connect with the nodes from other subgraphs.
Then, we define the federated graph as a union of subgraphs under a FL setting.

\begin{definition}\label{def:fg}
    \textbf{Federated Graph.} Federated graph is denoted as $\mathbb{G}=(\mathbb{V},\mathbb{E})=\{\mathbb{G}_s|s=1,2,\cdots,|\mathbb{G}|\}$, where $\mathbb{V}$ is the node set, $\mathbb{E}$ is the edge set, $\mathbb{G}_s=(\mathbb{V}_s,\mathbb{E}_s)$ is the $s$-th subgraph, and $|\mathbb{G}|$ is the number of subgraphs. $\mathbb{G}$ satisfies 
    1)~$\mathbb{V}_1 \cup \mathbb{V}_2 \cup \cdots \cup \mathbb{V}_{|\mathbb{G}|} = \mathbb{V}$, 
    2)~$\mathbb{E}_1 \cup \mathbb{E}_2 \cup \cdots \cup \mathbb{E}_{|\mathbb{G}|} \subseteq \mathbb{E}$, 
    3)~$\mathbb{V}_i \cap \mathbb{V}_j = \emptyset, \forall \ 1 \leq i \leq |\mathbb{G}|, 1 \leq j \leq |\mathbb{G}|, i\neq j$, 
    4)~$\mathbb{E}_i \cap \mathbb{E}_j = \emptyset, \forall \ 1 \leq i \leq |\mathbb{G}|, 1 \leq j \leq |\mathbb{G}|, i\neq j$.
\end{definition}

In this work, we associate each node with a device and each subgraph with a data silo. Under the cross-silo and cross-device setting, nodes and graph structures may be placed in multiple heterogeneous clients for federated graph learning.

\subsection{Problem Formulation}
In this work, we aim to collaboratively train GNNs over distributed graph data in a federated way, which is formally defined below.
\begin{problem}
    \textbf{Cross-silo Cross-device Federated Graph Learning.} Given a federared graph $\mathbb{G} = (\mathbb{V}, \mathbb{E})$ consisting of the subgraph set $\{\mathbb{G}_s\}_1^{|\mathbb{G}|}$. We aim to learn the parameter $\theta_s$ of $s$-th silo-client for the global GNN model $\mathcal{G}$ to minimize the global loss for downstream predictive or regressive tasks,
    \begin{equation}
        \{\theta_1, \cdots, \theta_{|\mathbb{G}|}\} = {\arg\min} \sum\mathcal{L}(\mathcal{G}(\mathbb{G}_s; \theta_s),Y_s),
    \end{equation}
    where $i=1, 2, \cdots, |\mathbb{G}|$ and $\mathcal{L}$ denotes the loss function between estimation $\mathcal{G}(\mathbb{G}_s; \theta_s)$ and ground truth $Y_s$.
\end{problem}

\subsection{Threat Model}\label{sec:threat}
Here we define the threat model for cross-silo cross-device FGL, which concurrently considers nodes' and subgraphs' privacy. Specifically, we assume all parties are \textit{semi-honest}, \ie honest-but-curious, which means the adversary will try its best to obtain private information but not break protocols or cause malicious damage to the model's ability. We first define nodes' and subgraphs' privacy as node-level privacy and subgraph-level privacy, respectively.
\begin{definition}
    \textbf{Node-level Privacy.} A node usually represents a device of a user in FGL applications, which should not leak the privacy of the raw features and adjacent neighbors to other participants, i.e., devices and data silos.
\end{definition}
\begin{definition}
    \textbf{Subgraph-level Privacy.} A subgraph corresponds to a data silo of an institution in FGL applications, which should not expose its subgraph to other data silos and devices.
\end{definition}

As illustrated in Example~\ref{exa:fin}, where banks correspond to subgraphs and accounts correspond to nodes, the corresponding participants in cross-silo cross-device FGL are inherently heterogeneous.
Then, we introduce the potential attack types between heterogeneous clients, including node-node attack, subgraph-node attack, and subgraph-subgraph attack.

\begin{definition}
    \textbf{Node-Node Attack.} A node acts as an attacker, and another node acts as a defender. Nodes only see their features thus they hope to obtain neighbor embeddings to infer more information.
\end{definition}
\begin{definition}
    \textbf{Subgraph-Node Attack.} A subgraph acts as an attacker, and a node acts as a defender. Subgraphs have no access to nodes' features and neighbors, and thus subgraphs are desired to utilize this information by cooperating with nodes. This cooperation may raise data leakage from nodes to subgraphs.
\end{definition}

\begin{definition}
    \textbf{Subgraph-Subgraph Attack.} A subgraph acts as an attacker, and another subgraph acts as a defender. The adversary hopes to acquire other subgraphs as well as cross-client edges to enhance their subgraph.
\end{definition}
Note we don't assume nodes will attack subgraphs. Due to the hierarchy of heterogenous clients, the device can access the model from their corresponding data silo.

Overall, in this work, we aim to preserve node-level privacy against node-node attacks and subgraph-node attacks, and preserve subgraph-level privacy against subgraph-subgraph attacks.

\section{Framework}

In this section, we present the proposed Hierarchical Federated Graph Learning~(HiFGL) framework in detail. In particular, we first present the hierarchical architecture for heterogeneous clients. Then, we introduce the Secret Message Passing~(SecMP) scheme to achieve multi-level privacy preservation. Finally, we detail the optimization scheme tailored for the framework.

\subsection{The Hierarchical Architecture}
\begin{figure}[t]
\setlength{\belowcaptionskip}{-0.3cm}
    \centering
    \includegraphics[width=\linewidth]{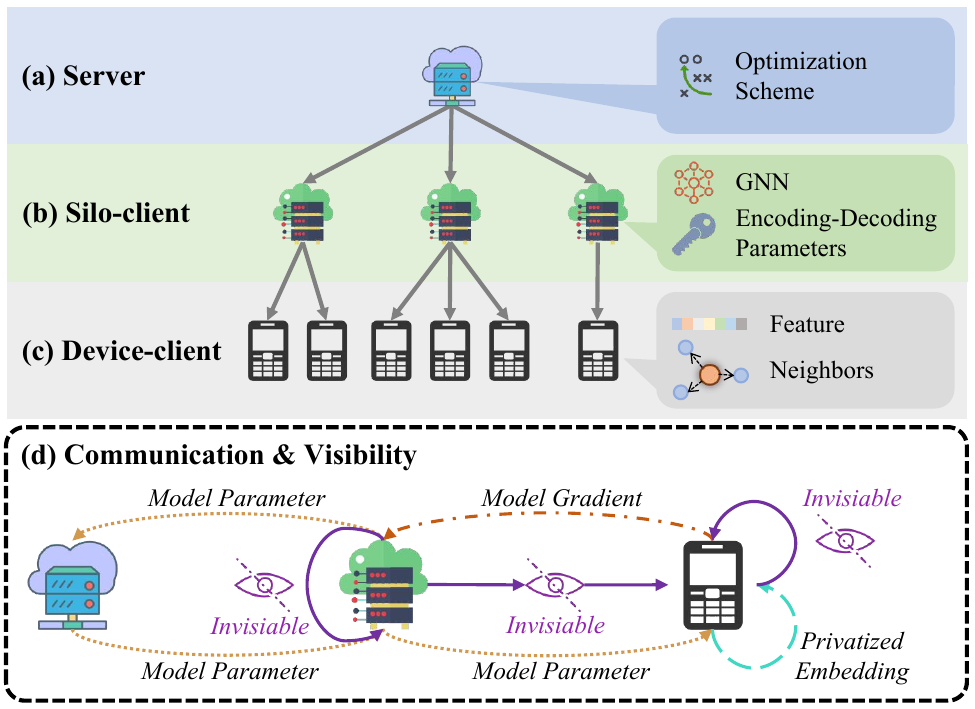}
    \caption{The architecture of the Hierarchical Federated Graph Learning framework.
    }
    \label{fig:framework}
    \Description{}
\end{figure}

We construct the hierarchical architecture of HiFGL consisting of three modules, i.e., device-client, silo-client, and server, in bottom-up order, as shown in Figure~\ref{fig:framework}.
Generally, under this hierarchy, modules are set to administrate the ones at a subordinate level and keep information private against the ones at a superordinate level\footnote{In this paper, we use ``the device-client's silo-client'' to denote ``the silo-client that administrates the device-client'' and ``the silo-client's device-client'' to denote ``the device-clients administrated by the silo-client''.}.
Specifically, the device-client holds local data and cooperates with its silo-client for learning without data exposure.
A silo-client preserves local models that are federally optimized with the server, administrates device-clients containing graph data, and determines the privatization method.
The server executes the federated optimization method for FL.
In the following parts, we detail three modules and their communication and visibility.

\subsubsection{Device-client}\label{sec:device}
The device-client has two roles, including storing the ego graph of each node and computing node gradients with the silo-client, as shown in Figure~\ref{fig:framework}~(c).

\textbf{1)~Storage.} A device-client stores an ego graph consisting of the target node's features and its neighbors.
\begin{definition}
    \textbf{Ego Graph.} An ego graph stores $j$-th node's feature $h^{(0)}_{j}$, label $y_j$, and $1$-hop directed neighbors. We denote the ego graph as $\tilde{g}_j=(h^{(0)}_{j},\mathcal{N}_{j})$.
\end{definition}
We store data separately in each device for two reasons.
First, storing data at the device level can prevent node features from being seen directly by adversaries at the silo or server level.
Second, cross-client edges are risky for exposing the silo's node identities. Suppose a directed cross-client edge connects a source and a target node from two silos. However, storing the edge in any one of the two silos will violate subgraph-level privacy requirements because there will be at least one node to be seen by a silo-client that is not its own. 
To address this challenge, we propose allowing two devices corresponding to the two nodes to have access to this cross-client edge in order to reduce the visibility of silos.

To construct ego graphs, we allow any two devices to connect in a peer-to-peer manner to ensure that the connected edges are only known by these two device-clients. For instance, two users can build a friend relationship without the awareness of the social network administrator. In the semi-honest setting, we consider the connected edges to be true. In this way, we can construct the graph in a privacy-preserving and decentralized way.

\textbf{2)~Gradients computation}
Considering the limited computing power and local data size, instead of optimizing the full parameters of the local model, the device-clients are only required to calculate gradients associated with their local data.
Specifically, for a model $\mathcal{G}$ parameterized by $\theta$, the $j$-th device-client computes the corresponding gradients by
$\nabla_j \theta = \nabla f_l(\mathcal{G}_\theta(h^{(0)}_j), y_j)$,
where $f_l(\cdot)$ is the loss function. These gradients are not considered as private information and can be shared with the corresponding silo-client.

\subsubsection{Silo-client}\label{sec:client}
As shown in Figure~\ref{fig:framework}~(b), the silo-client is a module between the server and the device-client, which is in charge of local model optimization based on data in its device-clients. We introduce two roles of each silo-client $C_i$, including storage and model update.

\textbf{1)~Storage.} The silo-client stores three elements, the subgraph, the local model, and encoding-decoding parameters.
First, $C_i$ indirectly keeps a federated subgraph $\tilde{G}_i$ consisting of federated ego graphs distributed in its device-clients $\mathbb{D}_i=\{D_j|j=1,2,\cdots\}$. We define the federated subgraph as below.
\begin{definition}
    \textbf{Federated Subgraph.} Federated subgraph is denoted as $\tilde{G}=\{D_j|j=1,2,\cdots\}$, where each $D_j$ stores a $\tilde{g}_j$, the $j$-th ego graph whose node $v_j$ is in subgraph $G$. $\tilde{G}$ only keeps the identities of the subordinate device-clients, but does not store any features and neighbor information.
\end{definition}
The federated subgraph avoids exposure of node features and graph structures and therefore protects device-level privacy in the federated learning process.

Moreover, since the silo-client usually has more powerful computing resources, we let $C_i$ keep the local GNN $\mathcal{G}_i$ for optimization.
Additionally, $C_i$ also preserves a set of individual encoding-decoding parameters $\mu_i$ generated by a polynomial function $\mathcal{F}$ for privatizing node embeddings.
Specifically, we define the parameters
$\mu=\{\alpha_1,\cdots,\alpha_{T+1},\beta_1,\cdots,\beta_{T+1},z_2,\cdots,z_{T+1}\}$,
where $\alpha_1,\cdots,\alpha_{T+1},$ $\beta_1,\cdots,\beta_{T+1}$ are $2T+2$ distinct elements from the finite field $\boldsymbol{F}$\footnote{Following \cite{yu2019lagrange}, we assume that $\boldsymbol{F}$ is a finite field with $11$ elements.}, satisfying $\{\alpha_1,\cdots,\alpha_{T+1}\}\cap\{\beta_1,\cdots,\beta_{T+1}\}=\varnothing$, and $z_2,\cdots,z_{T+1}$ are $T$ uniformly distributed vector.

\textbf{2)~Model update.} The silo-client $C_i$ is responsible to optimize $\mathcal{G}_i$ based on gradients computed by device-clients in $\mathbb{D}$. Specifically, $C_i$ collects the gradients set $\{\nabla_j \theta_i|D_j \in \mathbb{D}_i\}$ and compute the average gradients as $\nabla \theta_i = \frac{1}{|\mathbb{D}_i|}\sum_{D_j \in \mathbb{D}_i} \nabla_j \theta_i$, where $\theta_i$ is parameters of $\mathcal{G}_i$. The averaged gradients can be further utilized for optimization on $\mathcal{G}_i$. For example, we can leverage gradient descent as $\theta_i = \theta_i - \varrho \nabla \theta_i$ where $\varrho$ is the learning rate.

\subsubsection{Server}\label{sec:server}
The server coordinates the entire framework as shown in Figure~\ref{fig:framework}~(a), whose role is to enable federated optimization schemes (\eg FedAvg~\cite{mcmahan2017communication}) among silo-clients.
In practice, the server can be an administrator of silo-clients to supervise their sensitive activities in optimization. For instance, in a finance scenario, the server can be a banking authority~(\eg European Banking Authority), silo-clients are banks, and device-clients are different users. The federation is under the authority's control to prevent potential attacks among banks and users.

\subsubsection{Communication and Data Visibility}
The cross-level communication and data visibility among three different modules are shown in Figure~\ref{fig:framework}~(d). 
First, the server and silo-clients can transfer model parameters during federated optimization. Second, silo-clients are forbidden to exchange information, including device identities, models, and encoding-decoding parameters. 
Third, during the joint learning process, the device-client can access the model of the silo-client and send local gradients to the silo-client for model update, while the silo-client can not directly access the device-client's data. 
Last, device-clients communicate with each other through privatized embeddings without exposure of node features and neighborhood information.

\subsection{Secret Message Passing}\label{sec:secmp}
We further propose a novel graph learning method, SecMP, to preserve multi-level privacy without information loss. 
In particular, we develop the \textit{neighbor-agnostic aggregation} and \textit{hierarchical Lagrangian embedding} against potential subgraph-node and subgraph-subgraph attacks.
Specifically, the neighbor-agnostic aggregation decomposes the feature extraction function in conventional GNN into two individual steps to mutually cover neighbor information between any connected node pair, preventing subgraph-level sensitive information exchange. 
Meanwhile, the hierarchical Lagrangian embedding utilizes Lagrange polynomial functions to mask sharing node embeddings during message passing recoverably.

\subsubsection{Neighbor-agnostic aggregation.}
The aggregation step in GNNs may leak subgraph-level privacy, since such a process requires accessing structural information from cross-client neighbors~\cite{kipf2016semi,velivckovic2018graph,jin2020graph}. 
Take GCN~\cite{kipf2016semi} as an example, the aggregation process can be denoted by $h_{u} = \sum_{v \in \mathcal{N}_u}(1/(\sqrt{|\mathcal{N}_u|}\sqrt{|\mathcal{N}_v|})) h_v \mathbf{W}+b,$
where $h$ is node embedding, $b$ and $\mathbf{W}$ are learnable parameters, and $|\mathcal{N}|$ are the number of neighbors. 
Calculating $|\mathcal{N}_u|$ and $|\mathcal{N}_v|$ requires knowing precise numbers of $u$ and $v$, which is risky if the computation only inside either device-client $u$ or $v$.

To address this problem, we devise the neighbor-agnostic aggregation to split GNN aggregating operations into source-side and target-side steps, which makes the aggregation process agnostic with neighbors' information. Specifically, we operate a message passing from node $v$ to node $u$ in the source device-client and target device-client that correspond to two nodes, respectively. First, device-client $v$ will calculate a source-side function on $h_v$ to get $\hat{h}_v$ and pass it to device-client $u$. Then, device-client $u$ will calculate a target-side function to update its embedding. The source-side function can be defined as
\begin{equation}\label{equ:source_side}
  \hat{h}_v=(1/\sqrt{|\mathcal{N}_v|}) \cdot h_v,
\end{equation}
and the target-side function can be defined as
\begin{equation}\label{equ:target_side}
  h_u=b + (1/\sqrt{|\mathcal{N}_u|}) \sum\limits_{v \in \mathcal{N}_u}{\hat{h}_v \cdot \mathbf{W}}.
\end{equation}
In this way, we can preserve the privacy of both device-client $u$ and $v$'s neighbor lists by delegating operation-related private data to be computed within device-clients.

\subsubsection{Hierarchical Lagrangian embedding.}\label{sec:lcc}
Sharing embeddings with nodes' neighbors may induce severe node-level privacy leakage. Therefore, we privatize device-clients' information without information loss by introducing Lagrange Coded Computing~(LCC)~\cite{yu2019lagrange}, a secret sharing technique in GNN optimization. Specifically, our method follows an encoding-decoding workflow based on a group of parameters $\mu$ introduced in Section~\ref{sec:client}. In the encoding step, we use Lagrange interpolation polynomial to encode data to coded versions. In the decoding step, the aggregated coded vectors can be decoded without any information loss.

\textbf{Encoding.}
Any ego device-clients $D_s$ edged with a target device-client $D_t$ of silo-client $C$ will access $\mu$ and use them for constructing a polynomial function based on $D_s$'s embedding\footnote{The embedding can be computed by a source-side function as Equation~\ref{equ:source_side}.} $h_v$ as
\begin{equation}
\label{equ:lcc}
    g_v(x) = h_v \cdot \prod_{k \in[T+1]\backslash\{1\}}\frac{x-\beta_{k}}{\beta_{1}-\beta_{k}} + \sum\limits_{j=2}^{T+1} z_j \cdot \prod_{k \in[T+1]\backslash\{j\}}\frac{x-\beta_{k}}{\beta_{j}-\beta_{k}},
\end{equation}
$D_s$ generates $T+1$ coded embeddings $\tilde{h}_{v} = g_v(\alpha),\alpha\in{\alpha_1,\cdots,\alpha_{T+1}}$.

\textbf{Decoding.}
$D_t$ will receive $T+1$ coded embeddings\footnote{The embedding can be computed by a target-side function as Equation~\ref{equ:target_side}.} $h_1,\cdots,h_{T+1}$ from each source neighbor device-client $D_s\in{\mathcal{N}_{D_t}}$. $D_t$ will delegate the aggregated coded embedding to $C$ for decoding by calculating values of the Lagrange interpolated polynomial function at $x=\beta_1$.

\begin{figure}[t]
\setlength{\belowcaptionskip}{-0.3cm}
  \centering
  \includegraphics[width=\linewidth]{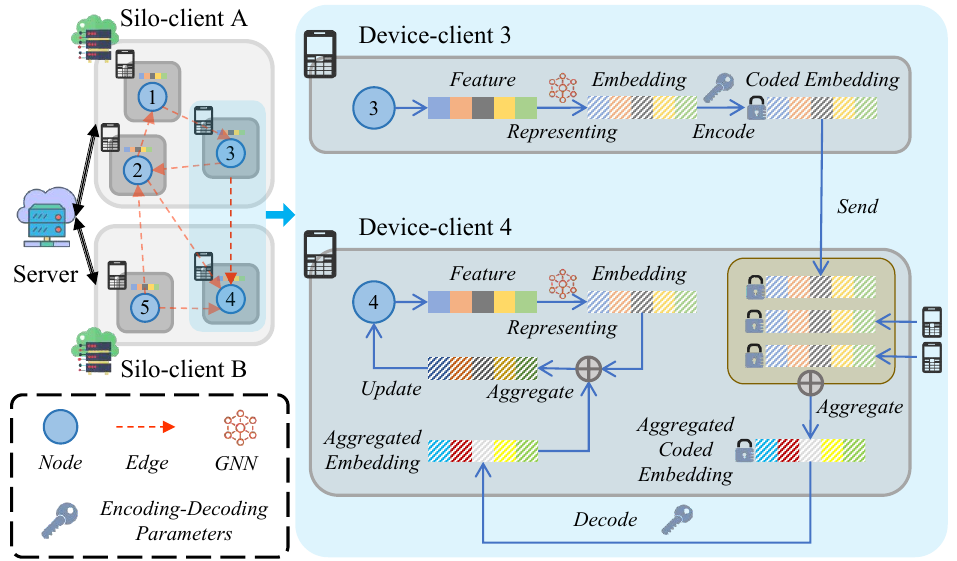}
  \caption{Overall workflow of Secret Message Passing.
  }
  \label{fig:secmp}
  \Description{}
\end{figure}

\subsubsection{Overall workflow}
We summarize the overall workflow of SecMP that incorporates neighbor-agnostic aggregation and hierarchical Lagrangian embedding for subgraph-level and node-level privacy protection. As illustrated in Figure~\ref{fig:secmp}, SecMP follows general GNN pipeline~\cite{JustinGilmer2017NeuralMP} in three major steps, \ie \textit{privatized message}, \textit{secure aggregation}, and \textit{neighbor-agnostic update}.

\textbf{1)~Privatized message.}
We first emit encoded embedding to the target device-client. In this way, device-clients are only aware of neighbors' coded embeddings instead of any concrete features. This step includes three sub-steps.
a)~Device-clients project raw features into node embeddings on the source-side.
b)~Device-clients request encoding-decoding parameters from neighbors' silo-clients. For each neighbor, the device-client will have one corresponding group of parameters for encoding.
c)~Device-clients encode embeddings and send encoded embeddings to both intra-client and cross-client target neighbor device-clients.

\textbf{2)~Secure aggregation.}
Then, the encoded embeddings from multiple neighbors are aggregated to derive a unified embedding. 
The silo-client can decode the aggregated embeddings without accessing the embeddings of each individual node, while device-clients cannot decode encrypted embeddings, therefore protecting individual embeddings. Two sub-steps of secure aggregation are elaborated below.
a)~Device-clients aggregate received encoded embeddings into a unified embedding.
b)~Device-clients send the aggregated embeddings to their silo-clients, which can use its encoding-decoding parameters to decode the aggregated embedding and send it back to device-clients.

\textbf{3)~Neighbor-agnostic update.} After secure aggregation, the target device-clients obtain the aggregated embeddings computed from the source side, and update their embeddings without requiring access to neighbors' structure information.

Note that an exceptional case is that when a node has only one neighbor, the corresponding device-client can get the node embedding after aggregation because the decoded aggregated embedding is exactly the embedding of the neighbor. A possible solution is to perturb embeddings by Differential Privacy~(DP) techniques~\cite{dwork2014algorithmic} to cover sensitive information. The DP-based masking operation also prevents attacks that infer original information by gradients or embeddings~\cite{zhu2019deep}. Besides, we build HiFGL on the Trusted Execution Environment~(TEE)~\cite{mo2021ppfl} and ensure the transition is anonymous for any receivers to eliminate exposure.

\subsection{Privacy-preserving Optimization}
Then, we present the privacy-preserving optimization algorithm for HiFGL. Without loss generality, we assume the local model is GCN for the node classification task, and the federated optimization algorithm is FedAvg.
The full pipeline is reported in Appendix~\ref{sec:appendix_training}.
We specify two critical steps in training, including local optimization and federated optimization.

\textbf{Local optimization.}
We optimize local models based on updated embeddings computed by SecMP. Specifically, we first calculate estimated classification probability as $\hat{y}=\operatorname{SoftMax}(h^{(k)})$ where $k$ is the index of the last GNN layer.
Then each device-client applies cross entropy loss on predicted labels $\hat{y}$ and true labels $y$ to calculate the loss value $l_j=\sum_q y_q\log \hat{y}_q$ and generate gradients $\nabla l_j$.
After that, gradients from device-clients will be transmitted to their silo-clients.
Lastly, silo-clients average them as gradients of this round as
$\nabla{\mathcal{L}_i} = (1/|\mathbb{D}_i|)\sum_{D_j \in \mathbb{D}_i}{\nabla{l_j}} = (1/|\mathbb{D}_i|)\sum_{D_j \in \mathbb{D}_i}{\nabla{\sum_q y_q\log \hat{y}_q}}$,
for local optimization on GNN parameters $\theta^{(n+1)}_i$.

\textbf{Federated optimization.}
When silo-clients finish their local optimization, the server will leverage FedAvg to average local models as the new global model as $\theta^{(n+1)} = (1/|\mathbb{C}|)\sum_{C_i \in \mathbb{C}}{\theta^{(n+1)}_i}$.
Updated parameters are the initial ones of the next round of optimization. Federated optimization is recurrently executed until the performance of each silo-client tends to converge.

\section{Analysis}

\subsection{Privacy Analysis}

\textbf{Subgraph-level privacy.}
In our solution, we preserve privacy by developing a hierarchical structure and a secure learning method. Specifically, for storage structure, edges are preserved between two device-clients corresponding to the two nodes instead of within the silo-clients, restraining features' exposure to other silo-clients. Besides, to protect against privacy leakage during message passing, the neighbor-agnostic strategy ensures that the target device-client only gets the encoded embedding and sends the aggregated one to its silo-client for decoding, where the silo-client cannot know the raw embedding of neighbors. In this way, $0\%$ subgraph-level privacy is leaked when executing SecMP according to the quantifying metric defined in Appendix~\ref{sec:appendix_subgraph_privacy}.

\noindent
\textbf{Node-level privacy.}
We analyze the privacy leakage of node embeddings $h$ of device-clients. Based on Equation~\ref{equ:lcc}, we select $\alpha_i \in {\alpha_1,\cdots,\alpha_{T+1}}$ to encode $h$ as $\tilde{h} = g(\alpha_i) = (h, z_2, \cdots, z_{T+1}) \cdot U_i$,
where $g(\beta_i) = z_i$ and $U \in{\boldsymbol{F}^{(T+1) \times (T+1)}}$ is the encoding matrix defined as
\begin{equation}
    U_{i,j} = \prod\limits_{k \in[T+1]\backslash\{i\}}\frac{\alpha_j-\beta_{k}}{\beta_i-\beta_{k}}.
\end{equation}
Any neighbor device-client will receive an encoded embedding $\tilde{h} = hU_t^{\text{top}}+ZU_t^{\text{bottom}},$ where $t\in{[T]}$, $Z = (z_2, \cdots, z_{T+1})$, and $U^{\text{top}}\in{\boldsymbol{F}^{T}}$, $U^{\text{bottom}}\in{\boldsymbol{F}^{T\times T}}$ are the $t$-th columns top and bottom submatrices in $U$. 
\begin{lemma}\label{lem:invertible}
The $T \times T$ matrix $U^{\text{bottom}}$ is invertible.
\end{lemma}
Please refer Appendix~\ref{sec:appendix_invertible} for Lemma~\ref{lem:invertible}. 
Since $U^{\text{bottom}}$ is invertible, we can mask encoded data $hU_t^{\text{top}}$ as $Z$ is uniformly randomized, and $h$ cannot be directly decoded for any $T \geq 1$, \ie $U_t^{\text{bottom}}$ exists.

\subsection{Complexity Analysis}

Here we analyze the communication, encoding and decoding, and space complexity of HiFGL.
We denote $\xi$ as the size of model parameters, $d$ as the dimension of coded or non-coded embeddings $h$ or $\tilde{h}$, $\gamma$ as the dimension of output prediction.

\emph{Communication complexity.}
We analyze three types of communication complexity guarantees of HiFGL. The communication complexity is $\mathcal{O}(dT)$ between a pair of connected device-clients. For a silo-client with its device-clients, the communication complexity for message passing and decoding is $\mathcal{O}(2|\mathbb{D}|\xi)$ and $\mathcal{O}(d|V_i| + \sum_{v_j \in V_i} dT|N_j|)$, respectively, where $|\mathbb{D}|$ is the number of the silo-client's device-clients. Last, the communication complexity between silo-clients and the server is $\mathcal{O}(2\xi)$ for each silo-client and $\mathcal{O}(2|\mathbb{C}|\xi)$ for the server. More details are provided in Appendix~\ref{sec:appendix_communication}.
\emph{Encoding and decoding complexity.}
The encoding and decoding complexity and be approximately guaranteed both as $\mathcal{O}(dT)$ according to analysis in Appendix~\ref{sec:appendix_enc_dec}.
\emph{Space complexity.}
As analyzed in Appendix~\ref{sec:appendix_space}, the space complexity of HiFGL is $\mathcal{O}(|\mathbb{G}| \cdot (\xi+3T+2) + |\mathbb{E}| + |\mathbb{H}|)$ and we guarantee the increased complexity $\delta_{\mathcal{S}}$ satisfying $\mathcal{O}(|\mathbb{G}|\xi) \leq \mathcal{O}(\Delta_{\mathcal{S}}) \leq \mathcal{O}(|\mathbb{G}|\xi + |\mathbb{E} \setminus (\mathbb{E}_1 \cup \mathbb{E}_2 \cup \cdots)|)$. 

\begin{table}[t]
\setlength{\abovecaptionskip}{0.1cm}
    \centering
    \caption{Overall information of datasets}
    \begin{tabular}{c|ccc}
        \toprule
         Datasets & Cora & CiteSeer & PubMed \\
         Split Method & Random & Random & Random \\
        \midrule
         $\#$ of Silo-clients & 5 & 5 & 5 \\
         $\#$ of Node & 542 & 665 & 3943 \\
         \makecell[c]{$\#$ of Intra-client Edges} & 431 & 183 & 1772 \\
         \makecell[c]{$\#$ of Cross-client Edges} & 4199 & 3637 & 35461 \\
         $\#$ of Classes & 7 & 6 & 3 \\
         Partition & 6/2/2 & 6/2/2 & 6/2/2 \\
         \bottomrule
    \end{tabular}
    \label{tab:dataset}
\end{table}

\begin{table*}[t]
\setlength{\abovecaptionskip}{0.1cm}
    \centering
    \caption{The prediction ACC and graph information gain of different FGL frameworks.}
    \label{tab:gain}
    \begin{tabular}{c|ccc|ccc|ccc}
        \toprule
        ACC & \multicolumn{3}{c|}{Cora} & \multicolumn{3}{c|}{CiteSeer} & \multicolumn{3}{c}{PubMed}\\
        \midrule
        & Mean & Std & Gain & Mean & Std & Gain & Mean & Std & Gain\\
        \midrule
        Local-MLP & 0.5698 & ±0.0071 & +0\% & 0.6450 & ±0.0061 & +0\% & 0.8051 & +0.0006 & +0\% \\
        \midrule
        Local-GCN & 0.8095 & ±0.0149 & +80.14\% & 0.7429 & ±0.0135 & +75.77\% & 0.8525 & ±0.0073 & +89.10\%\\
        FedAvg-GCN & 0.8358 & ±0.0135 & +88.93\% & 0.7601 & ±0.0152 & +89.09\% & 0.8603 & ±0.0095 & +103.76\%\\
        \textbf{HiFGL-GCN} & \textbf{0.8555} & \textbf{±0.0162} & \textbf{+95.52\%} & \textbf{0.7724} & \textbf{±0.0108} & \textbf{+98.61\%} & \textbf{0.8626} & \textbf{±0.0064} & \textbf{+108.08\%} \\
        Global-GCN & 0.8689 & ±0.0182 & +100\% & 0.7742 & ±0.0115 & +100\% & 0.8583 & ±0.0033 & +100\%\\
        \midrule
        Local-GraphSage & 0.6207 & ±0.0103 & +17.03\% & 0.6125 & ±0.0077 & -25.04\% & 0.8221 & ±0.0101 & +29.72\% \\
        FedAvg-GraphSage & 0.8095 & ±0.0123 & +80.22\% & 0.7656 & ±0.0139 & +92.91\% & 0.8444 & ±0.0098 & +68.71\% \\
        \textbf{HiFGL-GraphSage} & \textbf{0.8642} & \textbf{±0.0288} & \textbf{+98.53\%} & \textbf{0.7791} & \textbf{±0.0112} & \textbf{+103.31\%} & \textbf{0.8504} & \textbf{±0.0169} & \textbf{+79.20\%} \\
        Global-GraphSage & 0.8686 & ±0.0215 & +100\% & 0.7748 & ±0.0127 & +100\% & 0.8623 & ±0.0058 & +100\%\\
        \bottomrule
    \end{tabular}
\end{table*}

\section{Experiments}
\label{sec:exp}

\subsection{Experimental Setup}
Here we conduct experiments for evaluation on HiFGL.
We first introduce experimental setups including datasets, baselines, metrics, and implementation details.

\textbf{Datasets.}\label{sec:dataset}
We leverage popular graph datasets for node classification tasks, including Cora~\cite{sen2008collective}, CiteSeer~\cite{sen2008collective}, and PubMed~\cite{sen2008collective}. Following previous FGL benchmarks~\cite{federatedscope,ZhenWang2022FederatedScopeGNNTA}, we split the above graph datasets randomly to $5$ subgraphs with comparable node numbers. Details of datasets are presented in Table~\ref{tab:dataset}.

\textbf{Baselines.}
For framework-level experiments, we compare HiFGL with five baseline frameworks~(Local, FedAvg~\cite{mcmahan2017communication}, FedProx~\cite{li2020federated}, FedPer~\cite{arivazhagan2019federated}, and Global) incorporated with two popular GNNs variants~(GCN~\cite{kipf2016semi} and GraphSage~\cite{hamilton2017inductive}) as backbones. Besides, we also test performance of state-of-the-art FGL methods, including FedPerGNN~\cite{wu2022federated}, FedSage+~\cite{zhang2021subgraph}, and FED-PUB~\cite{baek2023personalized}. More information on baseline models is listed in Appendix~\ref{sec:appendix_baseline}.

\textbf{Metrics.}
We evaluate node classification performance by accuracy~(ACC), \ie the global percentage of accurately predicted samples, rather than the average of silo-client accuracy.
In particular, we design a new metric, \ie \textit{Graph Information Gain}, to show how much graph information has been modeled as
$\operatorname{Gain}(\operatorname{\star-}\mathcal{G}) = \frac{\operatorname{ACC}(\operatorname{\star-}\mathcal{G}) - \operatorname{ACC}(\operatorname{Local-MLP})}{\operatorname{ACC}(\operatorname{Global-}\mathcal{G}) - \operatorname{ACC}(\operatorname{Local-MLP})},$
where $\operatorname{\star-}\mathcal{G}$ denotes a GNN backbone model under a framework.

\textbf{Implementation details.}
We set the hidden dimension for GNNs and multi-layer perceptron as $64$, and the input and output dimensions depend on the raw feature dimensions and the number of classes for each dataset. We implement all GNNs with $2$ layers and train them for maximum $50$ epochs or two hours with Adam~\cite{kingma2014adam} optimizer where the learning rate is set as $0.01$ and multiplies gamma $0.9$ for every $4$ epoch. The $T$ in embedding privatization is set as $1$. The HiFGL framework is implemented based on PyTorch, and PyTorch-Lightning runs on the machine with Intel Xeon Gold 6148 @ 2.40GHz, V100 GPU and 64G memory.
Our codes are open-sourced at \url{https://github.com/usail-hkust/HiFGL}.

\subsection{Overall Prediction Performance}

In this subsection, we show the results of the node classification task including HiFGL and baseline frameworks and algorithms.
Specifically, we first compare HiFGL with three different frameworks to demonstrate the increased predictive ability brought by information integrity.
Second, we conduct experiments to illustrate that for FGL, information integrity is more important than any other algorithm-level improvement.

\subsubsection{Framework-level results.}
We investigate predictive knowledge retrieved under different frameworks measured by Graph Information Gain.
First, we define the predictability lower bound and upper bound as
\textit{1)~Lower bound~($0\%$)}: the performance of \textit{Local-MLP}, which only separately learns the knowledge from raw features through a $2$-layer multi-layer perceptron;
\textit{2)~Upper bound~($100\%$)}: the performance of trained $2$-layer backbone GNN~(GCN and GraphSage) on the \textit{Global} setting, which extracts both knowledge from raw features and graph information.
Then, we consider the gap between the lower and upper bound as graph information knowledge. Therefore, we test GNNs under different frameworks and collect the mean and standard deviation of ACC for five times experiments and graph information gain in Table~\ref{tab:gain}.
Results show that under HiFGL, GCN and GraphSage have more graph information gain over than Local and FedAvg and comparable with under Global. Specifically, on three datasets, HiFGL-GCN outperforms $6.59\%$, $9.52\%$, and $4.32\%$ than FedAvg-GCN, and HiFGL-GraphSage outperforms $18.31\%$, $10.40\%$, and $10.49\%$ than FedAvg-GraphSage. The model improvement demonstrates that besides ensuring privacy and complexity, HiFGL enhances information integrity to achieve better accuracy approximately equal ones without FL settings because the graph information gain of HiFGL is obtained from preserved cross-client edges, which offers rich relational knowledge to learn more effective embeddings.

\subsubsection{Algorithm-level results.}
The compared results among different FGL methods are depicted in Table~\ref{tab:overall}.
We compare HiFGL with three optimization schemes, \ie FedAvg, FedProx, and FedPer, with two GNNs, \ie GCN and GraphSage.
Besides, we involve FedPerGNN, FedSage+, and FED-PUB, which design tailored advanced graph learning modules for FGL.
We observe that under HiFGL, GCN, and GraphSage defeat all tested methods.
Specifically, with the integrated information preserved by the HiFGL, GCN achieves an accuracy of $0.8555$, over $2.3\%$, $3.8\%$, and $4.3\%$ improvement than FedAvg, FedProx, and FedPer on Cora dataset, respectively. Similarly, GraphSage gets $4.1\%$, $4.6\%$, and $5.7\%$ improvement. Significant performance promotion is attained on CiteSeer and PubMed datasets. The reason is that for GCN and GraphSage that highly depend on structure information, graph integrity can offer great benefits.
In addition, against methods with more advanced GNNs, HiFGL wins on prediction accuracy only by incorporating simple backbones. For example, under HiFGL, GraphSage surpasses FedPerGNN, FedSage+, and FED-PUB by $4.0\%$, $3.0\%$, and $2.8\%$, respectively, on CiteSeer. 
These experimental results demonstrate that graph integrity can bring much effectiveness improvement than GNN modification including high-order relationships incorporated in FedPerGNN, missing neighbor generation in FedSage+, and community-aware personalization in FED-PUB.

\begin{table}
\setlength{\abovecaptionskip}{0.1cm}
  \centering
  \caption{The prediction ACC of different FGL algorithms.}
  \label{tab:overall}
  \begin{tabular}{c|c|c|c}
    \toprule
    & Cora & CiteSeer & PubMed \\
    \midrule
    FedAvg-GCN & 0.8358 & 0.7601 & 0.8603 \\
    FedProx-GCN & 0.8238 & 0.7612 & 0.8305 \\
    FedPer-GCN & 0.8202 & 0.7403 & 0.8471 \\
    FedAvg-GraphSage & 0.8295 & 0.7540 & 0.8524 \\
    FedProx-GraphSage & 0.8256 & 0.7593 & 0.8391 \\
    FedPer-GraphSage & 0.8174 & 0.7625 & 0.8439 \\
    \midrule
    FedPerGNN & 0.8351 & 0.7488 & 0.8346 \\
    FedSage+ & 0.7767 & 0.7567 & 0.8394 \\
    FED-PUB & 0.8370 & 0.7579 & 0.8457 \\
    \midrule
    \textbf{HiFGL-GCN} & 0.8555 & 0.7724 & \textbf{0.8626} \\
    \textbf{HiFGL-GraphSage} & \textbf{0.8642} & \textbf{0.7791} & 0.8504 \\
    \bottomrule
  \end{tabular}
\end{table}

\subsection{Efficiency Results}

We study the efficiency by evaluating time and memory cost through varying two GNN hyper-parameters, \ie layer numbers and hidden dimensions, and take HiFGL-GCN as the testing model.

\textbf{GNN layers.}
First, we set GNN ranging from $1$ to $16$ layers, with a hidden dimension of $64$, on the CiteSeer dataset.
We display results in Table~\ref{tab:gnn_layer} that the highest ACC occurs in $2$-layer GCN, and the performance decreases for more and fewer layers.
The training time per epoch and the memory both increase along with the growth of the layer number. The results show that $2$-layer GCN in the HiFGL framework is the most efficient model for ACC and training time, and it does not need expensive memory for training.

\textbf{GNN dimension.}
We also set different hidden dimensions of a $2$-layer GCN, ranging from $4$ to $128$, on the PubMed dataset.
Table~\ref{tab:hidden_dim} demonstrates that performance increases with larger hidden dimensions since a more complicated model can catch more intricate patterns while obstructing high efficiency.
Besides, the memory for different hidden dimensions is only slightly different because parameters are not dominant for memory costs compared with the other parts of the HiFGL framework under our implementation.

In conclusion, higher performance is not always brought by expensive executive costs that multiple layers of GNNs need a heavy training burden but are predicted poorly. Increasing hidden dimensions of GNNs helps the ACC, but the time consumption grows exponentially. Therefore, we select $2$-layer GNNs with the hidden dimension of $64$ for our main experiments.
% \textbf{Discussion.}
Our framework has sacrificed efficiency to achieve more advanced privacy preservation, especially during training stages equipping with SecMP. The imperfection of inefficient processes remains a future development such as faster secret sharing techniques.
Fortunately, this work is still an applicable solution for cross-silo cross-device FGL scenarios with a high demand for privacy, scalability, and generalizability.

\begin{table}[t]
\setlength{\abovecaptionskip}{0.1cm}
    \centering
    \caption{Training results of different GNN layers.}
    \label{tab:gnn_layer}
    \begin{tabular}{c|ccccc}
        \toprule
        $\#$ of GNN Layers & 1 & 2 & 4 & 8 & 16 \\
        \midrule
        ACC~($\%$) & 73.04 & 77.91 & 70.87 & 39.84 & 26.02 \\
        Epoch Time~(s) & 6.5 & 4.3 & 5.6 & 8.4 & 16.4 \\
        Memory~(GB) & 3.41 & 3.45 & 3.49 & 3.59 & 3.80 \\
        \bottomrule
    \end{tabular}
\end{table}

\begin{table}[t]
\setlength{\abovecaptionskip}{0.1cm}
    \caption{Training results of different hidden dimensions.}
    \label{tab:hidden_dim}
    \centering
    \begin{tabular}{c|cccccc}
        \toprule
        \footnotesize{Hidden Dimensions} & 4 & 8 & 16 & 32 & 64 & 128 \\
        \midrule
        ACC~($\%$) & 83.45 & 85.15 & 85.33 & 85.27 & 85.88 & 86.39 \\
        Epoch Time~($s$) & 17.2 & 18.05 & 20.81 & 19.84 & 23.49 & 44.30 \\
        Memory~(GB) & \multicolumn{6}{c}{$3.753\pm0.005$} \\
        \bottomrule
    \end{tabular}
\end{table}

\subsection{Local Prediction Performance}

\begin{figure}[t]
\setlength{\abovecaptionskip}{0.1cm}
\setlength{\belowcaptionskip}{-0.3cm}
    \centering
    \subfigure[Training ACC of silo-client models.] {\includegraphics[width=.48\linewidth]{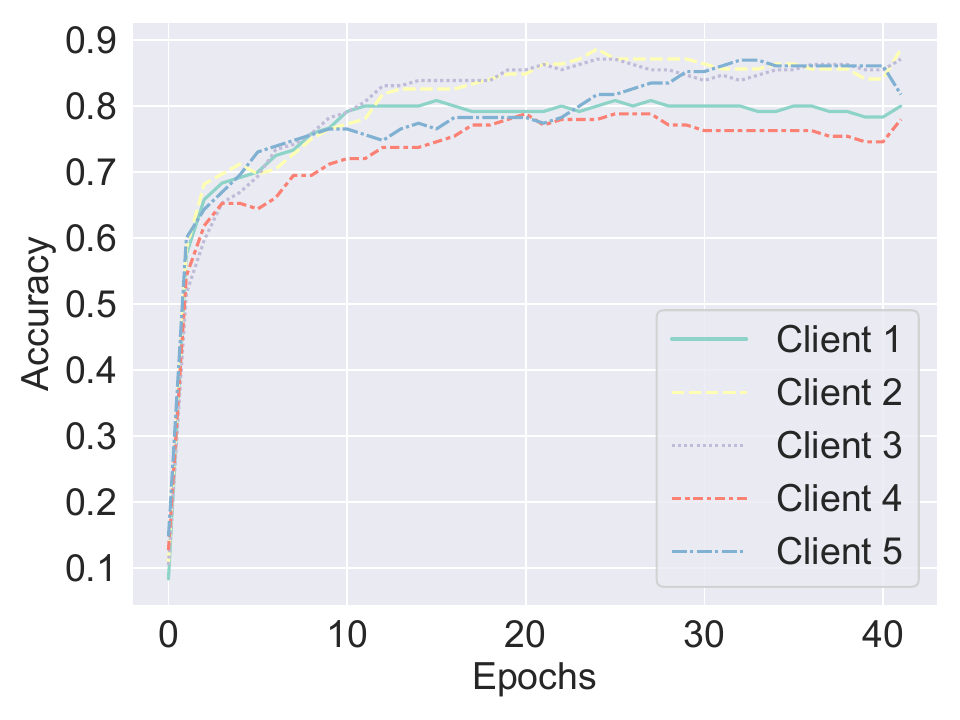} \label{fig:acc_curve}}
    \subfigure[Average ACC of silo-client models.] {\includegraphics[width=.48\linewidth]{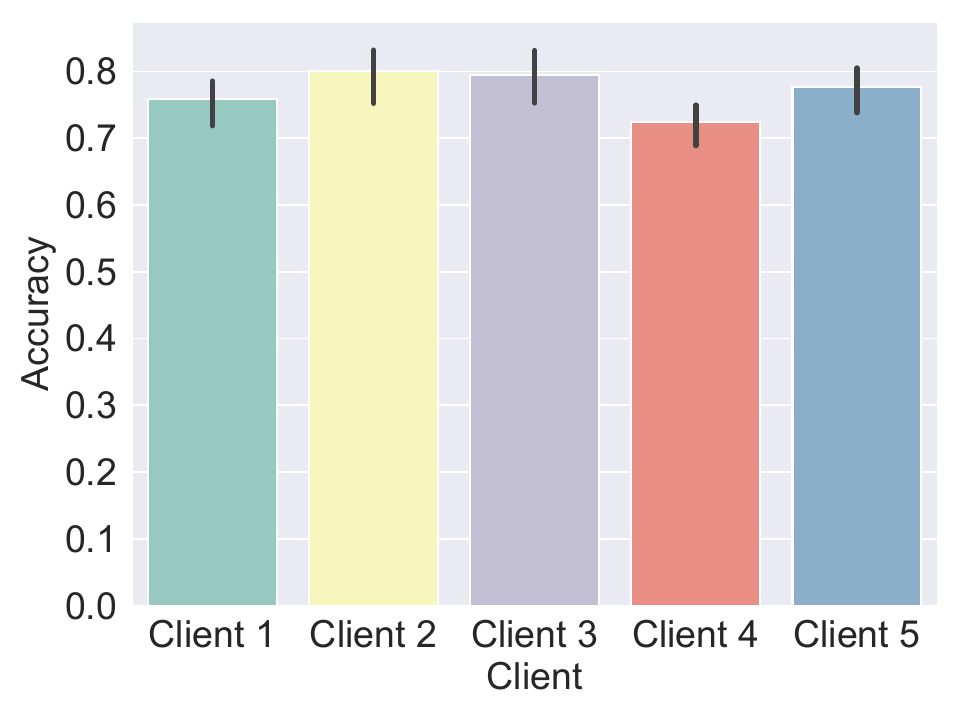} \label{fig:acc_bar}}
    \caption{Local model performance on Cora dataset.}
    \label{fig:local}
\end{figure}

Beyond the global perspective, we also investigate the local performance of each silo-client. Specifically, we collect the ACC of local models in all training epochs and show them in Figure~\ref{fig:local}, respectively. We discover that local models do not simultaneously upgrade or degrade during training in Figure~\ref{fig:acc_curve}. Instead, the five curves fluctuate in inconsistent upward patterns, which means a tradeoff is made among them during co-optimization directed by the server for global ACC improvement. Moreover, the converged performance of silo-clients is slightly diverse for their data distribution difference in Figure~\ref{fig:acc_bar}. The non-IID issue naturally exists in FL. In our example, the distributions of node features and graph structures in silo-clients are diverse naturally, which causes local models to be optimized in different directions during global loss degradation.

\begin{figure}[t]
\setlength{\abovecaptionskip}{0.1cm}
\setlength{\belowcaptionskip}{-0.3cm}
    \centering
    \subfigure[ACC] {\includegraphics[width=0.48\linewidth]{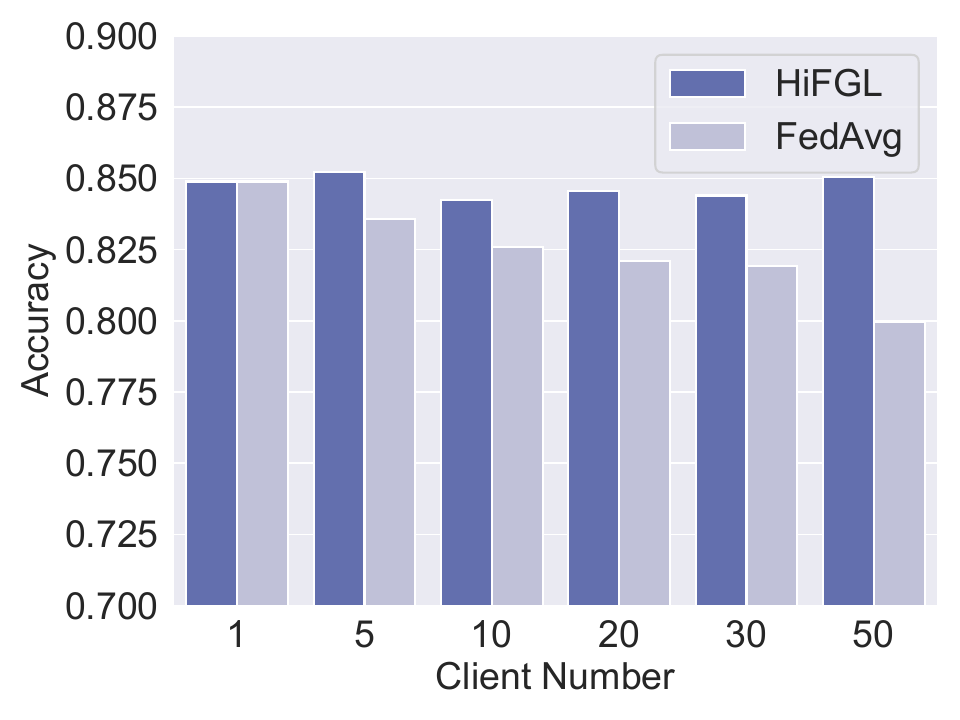}\label{fig:client_number_acc}}
    \subfigure[Epoch time] {\includegraphics[width=.48\linewidth]{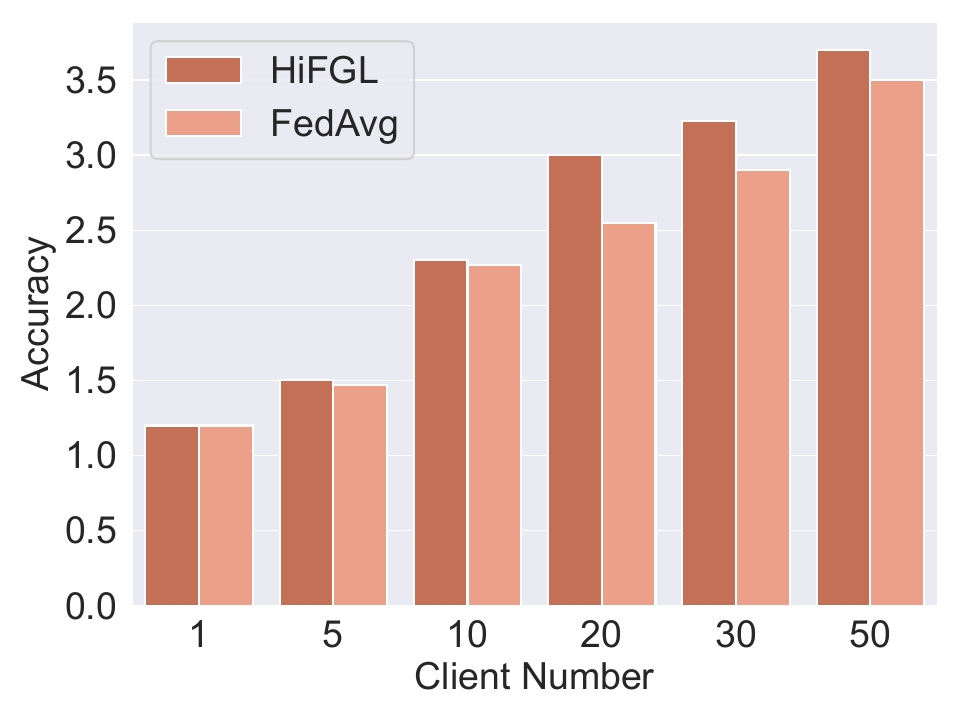} \label{fig:client_number_time}}
    \caption{Sensitivity for different silo-client numbers under the HiFGL and FedAvg frameworks.}
    \label{fig:client_number}
\end{figure}

\subsection{Sensitivity of Silo-client Number}

We evaluate the sensitivity of the silo-client number under different frameworks. Specifically, we vary it from $1$ to $50$ and take GraphSage as the backbone model to find out the variation of ACC in Figure~\ref{fig:client_number_acc}. Moreover, we also present executive time costs for different silo-client numbers in Figure~\ref{fig:client_number_time}. The number hardly influences model performance under HiFGL, while the ACC decreases as the number increases, and models of HiFGL are trained slightly longer than FedAvg. Because for FedAvg, a large number of silo-clients means that fewer time-consuming cross-client message passing operations are taken, and fragmentary information remaining on each silo-client is inadequate for precise training. These results further prove the importance of graph integrity for an FGL framework.

\section{Conclusion}
This paper first studies cross-silo cross-device FGL, where we propose a HiFGL framework based on a novel hierarchical architecture on heterogeneous clients.
Moreover, we devise a tailored graph learning algorithm, SecMP, for multi-level privacy-preserving optimization with graph integrity.
Theoretical analyses are provided for privacy and efficiency guarantees.
Extensive experiments demonstrate the prediction improvement gained from graph integrity on three datasets.
HiFGL offers versatility in a wider range of real-world applications, even in solely cross-silo or cross-device settings.

\begin{acks}
This work was supported by the National Natural Science Foundation of China (Grant No.62102110, No.92370204), National Key R\&D Program of China (Grant No.2023YFF0725004), Guangzhou-HKUST(GZ) Joint Funding Program (Grant No. 2022A03J00056, No.2023A03J0008), Education Bureau of Guangzhou Municipality.
\end{acks}

\clearpage
\bibliographystyle{ACM-Reference-Format}
\balance
\bibliography{reference}
\nocite{liu2022practical,zhang2021mugrep,he2018profiling,luo2020spatial}

\appendix
\section{Appendix}

\subsection{Quantifying Subgraph-level Privacy Leakage}\label{sec:appendix_subgraph_privacy}

We analyze the subgraph-level privacy leakage of previous cross-client FGL frameworks. Subgraph-level privacy leakage derives from the information sharing across cross-client edges, which lets silo-clients see nodes from others. There are works such as FedSage~\cite{zhang2021subgraph} and FedPUB~\cite{baek2023personalized} that drop cross-client edges to sacrifice the information integrity to perfectly preserve subgraph-level privacy. However, other works such as Glint~\cite{TaoLiu2021GlintDF} choose to permit silo-clients to access the features of the nodes from other silo-clients connected with their own nodes.
Here the subgraph-level privacy leakage of $i$-th silo-client $\epsilon_p^i$ can be defined as
\begin{equation}
    \epsilon_p^i = |\{\mathbb{D}_j|\mathbb{D}_j\in\mathbb{C}_i\ and\ \mathbb{D}^{\prime}\notin\mathbb{C}_i, \exists\mathbb{D}^{\prime}\in\mathcal{N}_j\}|,
\end{equation}
where we utilize the number of nodes that possess cross-client edges (\ie neighbors out of their own silo-clients). Subsequently, we compute the average of $\epsilon_p^i$ among silo-clients as the global privacy leakage by $\epsilon_p = \frac{1}{|\mathbb{C}|}\sum^{i=1}_{|\mathbb{C}|}\epsilon_p^i$.

Under this formulation, on experimental datasets introduced in Section~\ref{sec:dataset}, we observe that privacy leakage is remarkably significant, which is $95.16\%$ on Cora, $88.70\%$ on CiteSeer, and $90.31\%$ on PubMed for $5$ silo-clients. Otherwise, $90.69\%$, $95.21\%$, and $95.24\%$ edges are lost on Cora, CiteSeer, and PubMed for $5$ silo-clients, respectively, if we protect privacy by dropping all cross-client edges.

Therefore, preventing nodes from being seen by other silo-clients is necessary for FGL frameworks.

\subsection{Training Pipeline}\label{sec:appendix_training}

As an example, here we appoint GCN as the backbone model and FedAvg as the federated optimization method to explain the workflow in Algorithm~\ref{alg:secmp}. Specifically, after initialization of model parameters with $\theta^{(0)}$ and encoding-decoding parameters $\mu$, the optimization round will be executed recurrently. Training will be stopped after global convergence.

\begin{algorithm}[t]
  \caption{Training algorithm for GCN based on SecMP.}
  \label{alg:secmp}
  \LinesNumbered
  \KwIn{The graph $\mathbb{G}$, the server $S$, $|\mathbb{G}|$ silo-clients $\mathbb{C}$~(each $C_i$ with a local model $\mathcal{G}_{i}$), $n_d$ device-clients $\mathbb{D}$~(each $D_j$ with feature $h^{(0)}$ and neighbors $\mathcal{N}_{D_j}\}$).}
  \KwOut{The converged global model parameterized with $\theta^{(n)}$ trained after $n$ epochs $\mathcal{G}_{\theta^{(n)}}$.}
  Initialize each silo-client's model parameters with $\theta^{(0)}$ and encoding-decoding parameters $\mu=\{\alpha_1,\cdots,\alpha_{T+1},\beta_1,\cdots,\beta_{T+1},z_2,\cdots,z_{T+1}\}$\;
  \While{$\mathcal{G}_{\theta^{(n)}}$ is not converged}{
    \ForEach{$C_i \in \mathbb{C}$ in parallel}{
      \ForEach{$D_j \in \mathbb{D}_i$ in parallel}{
        $\hat{h}^{(0)} \gets \frac{1}{\sqrt{|\mathcal{N}_{j}|}}h^{(0)}$\;
        \ForEach{$D_p \in \mathcal{N}_{j}$}{
          Get $\mu_p$ from $D_p$'s silo-client\;
          $\tilde{h}^{(0)} \gets \mathcal{F}(\hat{h}^{(0)};\mu_p)$\;
          Send coded embedding $\tilde{h}^{(0)}$ to $D_p$\;
        }
        \tcc{After $D_j$ receives all coded embeddings $\tilde{\mathcal{H}}_{\mathcal{N}_{j}}$ from $\mathcal{N}_{j}$}
        $\tilde{h}^{(k)}_{\mathcal{N}_{j}} \gets \sum\limits_{\tilde{h}^{(0)}_p \in \tilde{\mathcal{H}}_{\mathcal{N}_{j}}}{\frac{1}{\sqrt{|\mathcal{N}_{j}|}}\mathcal{G}_i(\tilde{h}^{(0)}_p;\theta_i)}$\;
        Send $\tilde{h}^{(k)}_{\mathcal{N}_{j}}$ to $C_i$, and $C_i$ decodes it according to $\mathcal{F}$ and $\mu_i$ as $h^{(k)}_{\mathcal{N}_{j}}$ and send it back to $D_j$\;
        $h^{(k)} \gets h^{(k)}_{\mathcal{N}_{j}}$\;
        Calculate estimated probability as $\hat{y} \gets \operatorname{SoftMax}(h^{(k)})$\;
        Compute loss $l_j$ of prediction $\hat{y}$ and labels $y$\;
        Send gradient $\nabla{l_j}$ to $C_i$\;
      }
      Optimize $\theta^{(n+1)}_i$ with $\nabla{\mathcal{L}_i} \gets \frac{1}{|\mathbb{D}_i|}\sum\limits_{D_j \in \mathbb{D}_i}{\nabla{l_j}}$\;
    }
    Update global parameters as $\theta^{(n+1)} \gets \frac{1}{|\mathbb{C}|}\sum\limits_{C_i \in \mathbb{C}}{\theta^{(n+1)}_i}$\;
  }
\end{algorithm}

\subsection{Proof of Lemma~\ref{lem:invertible}}\label{sec:appendix_invertible}

\begin{proof}
Before we give proof, we revisit the selected $2T$ distinct elements which satisfying $\{\alpha_1,\cdots,\alpha_{T+1}\}\cap\{\beta_1,\cdots,\beta_{T+1}\}=\varnothing$.
Thus $\alpha_i - \alpha_j$, $\beta_i - \beta_j$, and $\alpha_i - \beta_j$ are non-zero for any different $i$ and $j$.
Here we start to validate $U^{\text{bottom}}$'s invertibility.
First, we multiply every $i$-th row by $\prod_{k\in{[T+1]}\backslash\{i\}}(\beta_i-\beta_k) \neq 0$ to obtain
\begin{equation}
    U^{\text{bottom}}_{i,j} = \prod\limits_{k \in[T+1]\backslash\{i\}}(\alpha_j-\beta_{k}).
\end{equation}
Second, we multiply every $j$-th column by $\prod_{k\in{[T+1]}}\frac{1}{\alpha_j-\beta_{k}} \neq 0$ as
\begin{equation}
    U^{\text{bottom}}_{i,j} = \frac{1}{\alpha_j-\beta_{i}},
\end{equation}
Third, we subtract the $n$-th column from the first $n-1$ columns and extract the common factor as
\begin{equation}
    \begin{array}{rl}
        U^{\text{bottom}}_{n\times n} & =
        \left[ \begin{smallmatrix}
            \frac{b_1-b_n}{(a_1-b_1)(a_1-b_n)}&\frac{b_2-b_n}{(a_1-b_2)(a_1-b_n)}&\cdots&\frac{1}{a_1-b_n}\\ \frac{b_1-b_n}{(a_2-b_1)(a_2-b_n)}&\frac{b_2-b_n}{(a_2-b_2)(a_2-b_n)}&\cdots&\frac{1}{a_2-b_n}\\ \frac{b_1-b_n}{(a_n-b_1)(a_n-b_n)}&\frac{b_2-b_n}{(a_n-b_2)(a_n-b_n)}&\cdots&\frac{1}{a_n-b_n}\\
        \end{smallmatrix}\right]\\
        & =\frac{\prod\limits_{i=1}^{n-1}(b_i-b_n)}{\prod\limits_{j=1}^{n}(a_j-b_n)}
        \left[ \begin{smallmatrix}
            \frac{1}{a_1-b_1}&\frac{1}{a_1-b_2}&\cdots&1\\
            \frac{1}{a_2-b_1}&\frac{1}{a_2-b_2}&\cdots&1\\
            \vdots&\vdots&\cdots&\vdots&\\
            \frac{1}{a_n-b_1}&\frac{1}{a_n-b_2}&\cdots&1\\
        \end{smallmatrix}\right]
    \end{array}
\end{equation}

Obviously, the values of the $n$-th column are the same, thus we can subtract the last row from the first $n-1$ rows and extract the common factor as
\begin{equation}
    \begin{array}{rl}
        U^{\text{bottom}}_{n\times n} & =\frac{\prod\limits_{i=1}^{n-1}(b_i-b_n)}{\prod\limits_{j=1}^{n}(a_j-b_n)}
        \left[ \begin{smallmatrix}
            \frac{a_n-a_1}{(a_1-b_1)(a_n-b_1)}&\frac{a_n-a_1}{(a_1-b_2)(a_n-b_2)}&\cdots&0\\
            \frac{a_n-a_2}{(a_2-b_1)(a_n-b_1)}&\frac{a_n-a_2}{(a_2-b_2)(a_n-b_2)}&\cdots&0\\
            \frac{1}{a_n-b_1}&\frac{1}{a_n-b_2}&\cdots&1\\
        \end{smallmatrix}\right]\\
        & =\frac{\prod\limits_{i=1}^{n-1}(a_n-a_i)(b_i-b_n)}{\prod\limits_{j=1}^{n}(a_j-b_n)\prod\limits_{k=1}^{n-1}(a_n-b_k)}
        \left[ \begin{smallmatrix}
            \frac{1}{a_1-b_1}&\frac{1}{a_1-b_2}&\cdots&\frac{1}{a_1-b_{n-1}}\\
            \frac{1}{a_2-b_1}&\frac{1}{a_2-b_2}&\cdots&\frac{1}{a_2-b_{n-1}}\\
            \vdots&\vdots&\cdots&\vdots&\\
            \frac{1}{a_{n-1}-b_1}&\frac{1}{a_{n-1}-b_2}&\cdots&\frac{1}{a_{n-1}-b_{n-1}}\\
        \end{smallmatrix}\right]
    \end{array}
\end{equation}
where we can obtain a recursion formula that
\begin{equation}
    U^{\text{bottom}}_{n\times n} = \frac{\prod_{i=1}^{n-1}(a_n-a_i)(b_i-b_n)}{\prod_{j=1}^{n}(a_j-b_n)\prod_{k=1}^{n-1}(a_n-b_k)}U^{\text{bottom}}_{(n-1)\times (n-1)}.
\end{equation}
Last, by continuing the recursive process, we can derive the determinant as
\begin{equation}
    \det{U^{\text{bottom}}_{n\times n}}=\frac{\prod_{1\leq i <j\leq n}^{}(a_j-a_i)(b_i-b_j)}{\prod_{i,j=1}^{n}(a_i-b_j)} \neq 0.
\end{equation}
Hence, the matrix $U^{\text{bottom}}$ is invertible.
\end{proof}

\subsection{Complexity Analysis Details}

Here we analyze communication, encoding and decoding, and space complexity, which are based on the following conditions.
The size of encoding-decoding parameters $\mu$ is $3T+2$. Proofs are satisfied with 1)~$T \leq 10$, 2)~$\gamma \leq 10$, 3)~$d >> T$, and $d >> \gamma$ based on realistic assumption and experimental datasets.

\subsubsection{Communication Complexity}\label{sec:appendix_communication}
The communication is composed of three parts, including 1)~\textit{communication between device-clients}: coded embedding exchange between device-clients and device-clients, 2)~\textit{communication between device-clients and silo-clients}: aggregated embedding communication between device-clients and silo-clients, and 3)~\textit{communication between silo-clients and the server}: model parameters sharing between silo-clients and the server.

1)~\textit{Communication between device-clients.} The communication between any two connected device-clients (from a source device-client to a target device-client) consists of two parts.
First, the target device-client passes the $3T+2$ encoding parameters of its silo-client to the source device-client.
Second, after coding by the source device-client, the coded embedding will be sent to the target device-client as $K$ parts by $T+1$ times at least.
Therefore, for any directed edge, the minimum communication complexity between two device-clients can be computed as $\mathcal{O}(3T+2 + d(T+1)) \approx \mathcal{O}(dT)$.

2)~\textit{Communication between device-clients and silo-clients.} Each device-client will download the latest model parameters and upload its gradients at each training round. Thus, the communication complexity between a single device-client and the corresponding silo-client can be measured by $\xi$, the size of the model parameters. In a word, the communication complexity for a device-client with its silo-client is $\mathcal{O}(2\xi)$, and for a silo-client with its device-clients is $\mathcal{O}(2|\mathbb{D}|\xi)$ that $|\mathbb{D}|$ is the number of the silo-client's device-clients.
In the decoding stage, when device-client $D_j$ transmits coded embeddings to the silo-client $C_i$ for decoding, the complexity is $\mathcal{O}(dT|N_j|)$ from $D_j$ to $C_i$ and $\mathcal{O}(d)$ from $C_i$ to $D_j$. Totally, the complexity of communication for $C_i$ is $\mathcal{O}(d|V_i| + \sum_{v_j \in V_i} dT|N_j|)$ at each round.

3)~\textit{Communication between silo-clients and the server.} The communication between silo-clients and the server obeys the traditional FL setting, which can be computed as $\mathcal{O}(2\xi)$ for each silo-client and $\mathcal{O}(2|\mathbb{C}|\xi)$ for the server.

\subsubsection{Encoding and Decoding Complexity}\label{sec:appendix_enc_dec}
We then present the analysis of Lagrange interpolation-based encoding and decoding complexity. Specifically, according to existing mathematical proof~\cite{kedlaya2011fast}, the complexity of interpolating a $k$-degree polynomial can be evaluated as $\mathcal{O}(k\log^2 k\log\log k)$. Therefore, we compute the encoding complexity as $\mathcal{O}(d(T+1)\log^2(T+1)\log\log(T+1)) \approx \mathcal{O}(dT)$. Besides, using the proposed technique~\cite{berlekamp2006nonbinary}, we can decode the embedding with a complexity as $\mathcal{O}(d(T+1)\log^2T\log\log T) \approx \mathcal{O}(dT)$.

\subsubsection{Space Complexity}\label{sec:appendix_space}
The space complexity $\mathcal{S}$ of HiFGL on a graph $\mathbb{G} = (\mathbb{V}, \mathbb{E}, \mathbb{H})$, is the summation of silo-client's model complexity $\mathcal{O}(|\mathbb{G}| \cdot (\xi+3T+2))$ and device-client's data (raw features and neighbor device-client pointers) complexity $\mathcal{O}(|\mathbb{E}| + |\mathbb{H}|)$, which can be formulated as
\begin{equation}
  \mathcal{S} = \mathcal{O}(|\mathbb{G}| \cdot (\xi+3T+2) + |\mathbb{E}| + |\mathbb{H}|),
\end{equation}
where $|\mathbb{G}|$ is the number of silo-clients~(or subgraphs), $|\mathbb{E}|$ is the number of edges, and $|\mathbb{H}|$ is the raw feature complexity.

Conventionally, in FGL, researchers proposed two schemes for graph storage. 1)~Trusted server for cross-client edges and silo-clients for intra-client edges. Its space complexity is $\mathcal{O}(|\mathbb{G}| \cdot \xi + |\mathbb{E}| + |\mathbb{H}|)$. 2)~Only silo-clients store intra-client edges. Its space complexity is $\mathcal{O}(|\mathbb{G}|\cdot\xi + \sum_{\mathbb{G}_i \in \{\mathbb{G}\}}{|\mathbb{E}_i|} + |\mathbb{H}|)$.
In this work, the HiFGL framework only needs tiny extra storage memory $\Delta_{\mathcal{S}}$ compared with two traditional schemes, which can be stated as
\begin{equation}
  \mathcal{O}(|\mathbb{G}|\xi) \leq \mathcal{O}(\Delta_{\mathcal{S}}) \leq \mathcal{O}(|\mathbb{G}|\xi + |\mathbb{E} \setminus (\mathbb{E}_1 \cup \mathbb{E}_2 \cup \cdots)|),
\end{equation}
where $|\mathbb{E} \setminus (\mathbb{E}_1 \cup \mathbb{E}_2 \cup \cdots)| \leq |\mathbb{E}|$.

\subsection{Baseline Information}\label{sec:appendix_baseline}
We involve five baseline frameworks in our experiments, including
\begin{itemize}
    \item \textit{Local}: models are individually trained on silo-clients.
    \item \textit{FedAvg}~\cite{mcmahan2017communication}: a collaborative learning framework that averages local model parameters for global optimization.
    \item FedProx~\cite{li2020federated}: a more robust federated optimization method based on FedAvg with regularization of parameters.
    \item \textit{FedPer}~\cite{arivazhagan2019federated}: a personalized FL algorithm allowing some layers to be free from FedAvg training for better local fitness.
    \item \textit{Global}: a usual graph training way without distributed or private constraints.
\end{itemize}
Furthermore, we compare two popular GNNs as backbone models within federated frameworks, including
\begin{itemize}
    \item GCN~\cite{kipf2016semi}: a spectral-based graph convolutional network form capturing first-order structure feature with node information for node representation learning.
    \item GraphSage~\cite{hamilton2017inductive}: a spatial-based graph model aggregating sampled neighbor information with ego features for inductive graph learning.
\end{itemize}
We also leverage $2$-layer multi-layer perceptron~(MLP) under \textit{Local} frameworks for comparison with GNNs.
Besides, we test classification performance on state-of-the-art FGL methods, including
\begin{itemize}
    \item FedPerGNN~\cite{wu2022federated}: a cross-device FL method on personalized federated item recommendation.
    \item FedSage+~\cite{zhang2021subgraph}: a federated inductive graph learning model with neighbor generation for missing edge reconstruction.
    \item FED-PUB~\cite{baek2023personalized}: a personalized federated subgraph learning method incorporating community structures and masked graph convolutional networks.
\end{itemize}

\end{document}